# Different Operating Systems Compatiblefor Image Prepress Process in Color Management: Analysis and Performance Testing

Jaswinder Singh Dilawari[1]and Dr.Ravinder Khanna[2]

[1] Ph.D Research Scholar, Pacific Academy of Higher Education and Research University, Udaipur, Rajasthan, INDIA
dilawari.jaswinder@gmail.com

[2]Principal, Sachdeva Engineering College for Girls, Gharuan, Mohali, Punjab, INDIA
ravikh_2006@yahoo.com

## Abstract

*Image computing has become a real catchphrase over the past few years and the interpretations of the meaning of the term vary greatly. The Imagecomputing market is currently rapidly evolving with high growth prospects and almost daily announcements of new devices and application platforms, which results in an increasing diversification of devices, operating system and development platforms. Compared to more traditional information technology markets like the one of desktop computing, mobile computing is much less consolidated and neither standards nor even industry standards have yet been established. There are various platforms and interfaces which may be used to perform the desired tasks through the device. We have tried to compare the various mobile operating systems and their trade-offs.*

## 1. INTRODUCTION

**A** key aspect of Image Processors is that they possess a complex operating system. This is a major difference that sets them apart from the class of feature image processors [1]. A feature phone is a regular mobile phone with feature support such as e.g. a high-resolution display, a built-in camera or a mp3-player. Feature phones are closed devices that do not offer the extensibility through a native programming API. The only way of customization sometimes offered by these devices is through Java ME or via web applications. Image computing is generally referred to smartphones, which represent intelligent phone centric handheld devices that can be leveraged by third-party applications [2]. Image Processors are believed by many to be the enabling device for mobile computing, similar to what the IBM PC was for desktop and office computing.

## 2. OPERATING SYSTEMS IN IMAGE PROCESSORS

**A**longside the myriad of mobile devices there exists also a huge number of operating systems and even more application development platforms. Like a computer operating system, a mobile operating system is thesoftware platform on top of which other programs run.

When you purchase a mobile device, the manufacturer will have chosen the operating system for that specific device. The operating system is responsible for determining the functions and features available on your device, such as thumbwheel, keyboards, WAP, synchronization with applications, e-mail, text messaging and more. The mobile operating system will also determine which third-party applications can be used on your device.

### 2.1 Java Platform

Java Platform, Micro Edition (Java ME, formerly J2ME) is not a mobile operating system instead it is middle layer between a specific mobile operating system and value added services and applications offered by a service provider [2]. The Java Platform is divided into three main products, each one targeted at a special field of application [3]:

### 2.2 Java Platform, Standard Edition (Java SE, formerlyJ2SE)

Java SE serves as the standard Java Edition for developing desktop and small server applications. It provides a rich set of APIs for all various kinds of software development needs, from basic String manipulation over graphics and user interface (UI) creation to networking. It is the basis for the Enterprise Edition(Java EE)

### 2.3 Java Platform, Enterprise Edition (Java EE, formerly J2EE)

Java EE extends the Standard Edition by adding a set of libraries for creating full featured web- & server-applications. It offers e.g. transaction support, libraries for mapping objects to relationale databases (O/R-mapping) and Web Services. It is specially known for its Servlet- Technology, which allows for using Java's rich libraries within web applications.

### 2.4 Java Platform, Micro Edition (Java ME, formerly J2ME)

Java Micro Edition is specially aimed for the development of applications on small, limited devices like mobile phones, PDAs, set-top boxes and so forth. It utilizes only a subset of the APIs available with the Standard Edition and builds on a smaller Virtual Machine (VM) for meeting the resource scarcity of its target devices.

## 3. JAVA TECHNOLOGY FOR THE WIRELESS INDUSTRY (JTWI)

The fragmentation of the Java ME application framework starts with the distinction of the Connected Device Configuration (CDC) and the Connected Limited Device Configuration (CLDC). Within the more prevalent stack of CLDC there exist currently two different versions of configurations (CLDC 1.0 & CLDC 1.1) as well as two quite different profile specifications MIDP 1.0 & MIDP 2.0. In response to the quickly evolving mobile device hardware market, additional APIs have been introduced to allow developers to exploit the new handset features. These APIs have been amongst others Bluetooth (JSR 82), 3D Graphics (JSR 184), Wireless Messaging (JSR 205), Web Services (JSR 172), and so forth. This process furthermore lead to a fragmentation of the market of Java ME enabled mobile devices and made it particularly difficult for the developers to rely on a basic set of functionality.

## 4. SYMBIAN - SYMBIAN OS

**S**ymbian OS is one of the most popular and widespread smartphone operating systems. The Symbian OS is based on a microkernel design, meaning that only a small set of system function resides in the kernel. Originally this kernel was not a real-time kernel but since version 9 Symbian OS is a real-time operating system [4]. Real-time systems have rigid time constraints and a system failure occurs if they are not met by an application. As the minimalistic microkernel of Symbian OS only provides the most basic functionalities like memory management, device drivers or power management, much of the remaining functions usually found in a monolithic kernel is provided by servers. Servers are responsible for all kinds of low level tasks like socket connections, file handling, telephony, etc. Clients are the final building block in the basic Symbian OS architecture. The applications that involve some kind of user interaction via the User Interface (UI). Together client, server and the kernel form a client/server architecture in which clients communicate with the servers via a message passing protocol (inter-process communication) and servers making executive calls into the kernel when necessary. Symbian OS has become a standard operating system for smartphones, and is licensed by more than 85 percent of the world's handset manufacturers. The Symbian OS is designed for the specific requirements of 2.5G and 3G mobile phones.

### 4.1 Symbian C++

**A**s the Symbian OS[4] is itself written in C++, it is therefore considered the primary programming language. As a typical representative of an operating systems native language, C++ offers the greatest possibilities and best performance regarding memory footprint and execution speed. C++ offers full access to every exposed library and is required to be used for the development of servers, plug-ins that extend a framework and device drivers that interact with the kernel. The logical consequence of using Symbian C++ is of course a tight coupling to the Symbian operating system.

### 4.2 Java in Symbian OS

**J**ava is an integral part of the Symbian operating system architecture. Java integration is based on standard Java MEwith the CLDC 1.1 and MIDP 2.0. MIDP 2.0 is available since Symbian version 7.0 and the CLDC 1.0 from earlier realeses was superseded by the CLDC 1.1 with version 8 of the OS. According to the official operating system guide14, the following optional JSRs are currently supported by Symbian OS:

**4.2.1 JSR 82: Java APIs for Bluetooth_v7.0s (Bluetooth Push was added in v8.0).**

**4.2.2 JSR 120: Wireless Messaging API_ v7.0s.**

**4.2.3 JSR 185: Java Technology For The Wireless Industry (JTWI) _ v7.0s (this was developed for v8.0, but was backported to v7.0s).**

**4.2.4 JSR 139: CLDC 1.1 _ v8.0.**

**4.2.5 JSR 75: FileConnection Optional Package _ v8.0.**

**4.2.6 JSR 135: Mobile Media API _ v8.0.**

**4.2.7  JSR 75: PIM Optional Package _ v8.1.**

## 5. MICROSOFT - WINDOWS MOBILE

**T**he Windows Mobile family of operating systems is Microsoft's contribution to the world of small mobile devices. The most current version is Windows Mobile 6, which comes in three flavors:

### 5.1 Windows Mobile 6 Classic

(Formerly: Windows Mobile for Pocket PC)

### 5.2 Windows Mobile 6 Professional

(Formerly: Windows Mobile for Pocket PC Phone Edition)

### 5.3 Windows Mobile 6 Standard

(Formerly: Windows Mobile for Smartphone)

According to Microsoft's Developer Network MSDN15 the product names have been changed to better reflect the realities of today's mobile device marketplace, where former distinctions between different classes of devices blur rapidly and the smartphone is becoming the universal mobile handset. The underlying operating system technologies and the APIs are consistent across all Windows Mobile devices. The main difference lies inthe different form factors of the target devices, e.g. screen resolution and input facilities (QWERTY keyboard, touchscreen). Generally an application developed for a particular flavor of Windows Mobile should work across all Windows Mobile devices.

## 6. CONCLUSION

The operating systems that offer developers the most choice of development environment are Windows Mobile and Symbian. In many operating systems Java plays a very pivotal role in automating the device as per the increased demand of functionality in matching with today's market trends and setting up new standards. Perhaps more importantly, it's clear that a large portion of smartphone operating systems can be targeted by using either Flash Lite or Java ME (and no complaining about Java ME fidelity/incompatibility issues. To decide which one is better we tested some features on the various operating systems:

### 6.1 Start-up Time (Booting)

We found that Symbian takes more time to complete the boot process compare to Windows Mobile. This is because Windows Mobile smart phone usually equipped with higher RAM compare to Symbian smart phone. So it can perform the booting process faster.

### 6.2 Contacts List

For contacts list, Symbian performs better than Windows Mobile. Nowadays, one person might use more than one mobile phone, so they going to have more than one mobile phone numbers. Windows Mobile can only store one number for each category (e.g.: mobile, work, home, office). Let say person A have 2 mobile phone numbers, in Windows Mobile we have no choice but to store one number as mobile number and another one as another category for example home. But in Symbian, we can store both numbers as mobile number.

### 6.3 Java Applications

Running Java applications using MIDletManager[8] in Windows Mobile can sometime cause the device hang. The start-up process of java application in Windows Mobile[6] is also slow. But it is different in Symbian. Java applications start and run smoothly in Symbian smart phone.

### 6.4 Dialling or Search Contacts List

Dialling or searching contact in contacts list is easier to be done in Windows Mobile. This is because Windows Mobile provides suggestion of person to contact every time we type something in the search area. For example we have aperson Raman Kumar Sharma stored in the contact list, and we search for "Sharma" in the search area.

Windows Mobile will come out with suggestion that match the keyword and surprisingly Raman Kumar Sharma will displayed which makes the process of searching faster and user friendly.

### 6.5 SMS

Report delivery in Symbian is better than Windows Mobile[7]. Report is receive in notification from and not as a message as in Windows Mobile. The report also clearly states the contact person's name while in Windows Mobile it only shows the contact person's number.

### 6.6 Browser

The default browser in Windows Mobile is much better than default browser in Symbian in term of usability and accessibility. Internet Explorer shows the better results as compared to the other operating systems in terms of browsing and surfing.

The mobile operating system market is currently subject to rapid changes and many competitors are fighting fiercely for market shares in this promising area of computing. As of today, it is extremely hard if not impossible to predict which platform will eventually prevail. The mobile market is still in its infancy, which means that new platforms may have big impacts on the overall market situation and market shares are subject to much faster changes than in more traditional markets like that of desktop computers.

**Authors:**

**Jaswinder Singh Dilawari** is a Research Scholar, Pacific Academy of Higher Education and Research University, Udaipur, Rajasthan, INDIA and is working as an Associate Professor ,Dept of Computer Sci and Engineering ,Geeta Engineering College, Panipat, Haryana ,India .He has teaching experience of 12 years .His area of interest includes Computer Graphics, Computer Architecture ,Software Engineering ,Fuzzy Logic  and Artificial Intelligence .He is life member of  Indian Society for Technical Education (ISTE)

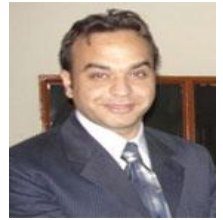

Born in 1948, **Dr. RavinderKhanna** Graduated in Electrical Engineering from Indian Institute of Technology(IIT) Dehli in 1970 and Completed his Masters and Ph.D degree in Electronics and Communications Engineering from the same Institute in 1981 and 1990respectively. He worked as an Electronics Engineer in Indian Defense Forces for 24Years where he was involved in teaching, research and project management of some of the high tech weapon systems. Since 1996 he has full time Switched to academics. he has worked in many premiere technical institute in India and abroad. Currently he is the Principal of Sachdeva Engineering College for Girls, Mohali, Punjab (India).He is active in the general area of Computer Networks, Image Processing and Natural Language Processing

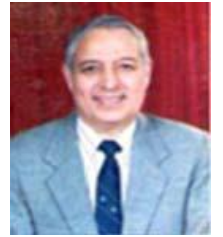